# SPUDD: Stochastic Planning using Decision Diagrams

Jesse Hoey    Robert St-Aubin    Alan Hu    Craig Boutilier

Department of Computer Science
University of British Columbia
Vancouver, BC, V6T 1Z4, CANADA
{jhoey,staubin,ajh,cebly}@cs.ubc.ca

## Abstract

Recently, structured methods for solving factored Markov decisions processes (MDPs) with large state spaces have been proposed recently to allow dynamic programming to be applied without the need for complete state enumeration. We propose and examine a new value iteration algorithm for MDPs that uses algebraic decision diagrams (ADDs) to represent value functions and policies, assuming an ADD input representation of the MDP. Dynamic programming is implemented via ADD manipulation. We demonstrate our method on a class of large MDPs (up to 63 million states) and show that significant gains can be had when compared to tree-structured representations (with up to a thirty-fold reduction in the number of nodes required to represent optimal value functions).

## 1 Introduction

Markov decision processes (MDPs) have become the semantic model of choice for decision theoretic planning (DTP) in the AI planning community. While classical computational methods for solving MDPs, such as value iteration and policy iteration [19], are often effective for small problems, typical AI planning problems fall prey to Bellman's *curse of dimensionality*: the size of the state space grows exponentially with the number of domain features. Thus, classical dynamic programming, which requires explicit enumeration of the state space, is typically infeasible for feature-based planning problems.

Considerable effort has been devoted to developing representational and computational methods for MDPs that obviate the need to enumerate the state space [5]. *Aggregation* methods do this by aggregating a set of states and treating the states within any aggregate state as if they were identical [3]. Within AI, *abstraction* techniques have been widely studied as a form of aggregation, where states are (implicitly) grouped by ignoring certain problem variables [14, 7, 12]. These methods automatically generate abstract MDPs by exploiting structured representations, such as probabilistic STRIPS rules [16] or *dynamic Bayesian network* (DBN) representations of actions [13, 7].

In this paper, we describe a dynamic abstraction method for solving MDPs using *algebraic decision diagrams* (ADDs) [1] to represent value functions and policies. ADDs are generalizations of ordered *binary decision diagrams* (BDDs) [10] that allow non-boolean labels at terminal nodes. This representational technique allows one to describe a value function (or policy) as a function of the variables describing the domain rather than in the classical "tabular" way. The decision graph used to represent this function is often extremely compact, implicitly grouping together states that agree on value at different points in the dynamic programming computation. As such, the number of expected value computations and maximizations required by dynamic programming are greatly reduced.

The algorithm described here derives from the *structured policy iteration (SPI)* algorithm of [7, 6, 4], where decision trees are used to represent value functions and policies. Given a DBN action representation (with decision trees used to represent conditional probability tables) and a decision tree representation of the reward function, SPI constructs value functions that preserve much of the DBN structure. Unfortunately, decision trees cannot compactly represent certain types of value functions, especially those that involve disjunctive value assessments. For instance, if the proposition $a \lor b \lor c$ describes a group of states that have a specific value, a decision tree must duplicate that value three times (and in SPI the value is computed three times). Furthermore, if the proposition describes not a single value, but rather identical subtrees involving other variables, the entire subtrees must be duplicated. Decision graphs offer the advantage that identical subtrees can be merged into one. As we demonstrate in this paper, this offers considerable computational advantages in certain natural classes of problems. In addition, highly optimized ADD manipulation software can be used in the implementation of value iteration.

The remainder of the paper is organized as follows. We provide a cursory review of MDPs and value iteration in Section 2. In Section 3, we review ADDs and describe our ADD representation of MDPs. In Section 4, we describe a conceptually straightforward version of SPUDD, a value iteration algorithm that uses an ADD value function representation, and describe the key differences with the SPI algorithm. We also describe several optimizations that reduce both the time and memory requirements of SPUDD. Empir-ical results on a class of process planning examples are described in Section 5. We are able to solve some very large MDPs exactly (up to 63 million states) and we show that the ADD value function representation is considerably smaller than the corresponding decision tree in most instances. This illustrates that natural problems often have the type of disjunctive structure that can be exploited by decision graph representations. We conclude in Section 6 with a discussion of future work in using ADDs for DTP.

## 2  Markov Decision Processes

We assume that the domain of interest can be modeled as a fully-observable MDP [2, 19] with a finite set of states $\mathcal{S}$ and actions $\mathcal{A}$. Actions induce stochastic state transitions, with $\Pr(s, a, t)$ denoting the probability with which state $t$ is reached when action $a$ is executed at state $s$. We also assume a real-valued reward function $R$, associating with each state $s$ its immediate utility $R(s)$.[1]

A *stationary policy* $\pi : \mathcal{S} \to \mathcal{A}$ describes a particular course of action to be adopted by an agent, with $\pi(s)$ denoting the action to be taken in state $s$. We assume that the agent acts indefinitely (an infinite horizon). We compare different policies by adopting an *expected total discounted reward* as our optimality criterion wherein future rewards are discounted at a rate $0 \le \beta < 1$, and the value of a policy is given by the expected total discounted reward accrued. The expected value $V_\pi(s)$ of a policy $\pi$ at a given state $s$ satisfies [19]:

$$V_\pi(s) = R(s) + \beta \sum_{t \in \mathcal{S}} Pr(s, \pi(s), t) \cdot V_\pi(t) \qquad (1)$$

A policy $\pi$ is *optimal* if $V_\pi \ge V_{\pi'}$ for all $s \in \mathcal{S}$ and policies $\pi'$. The *optimal value function* $V^*$ is the value of any optimal policy.

*Value iteration* [2] is a simple iterative approximation algorithm for constructing optimal policies. It proceeds by constructing a series of $n$-stage-to-go value functions $V^n$. Setting $V^0 = R$, we define

$$V^{n+1}(s) = R(s) + \max_{a \in \mathcal{A}} \left\{ \beta \sum_{t \in \mathcal{S}} Pr(s, a, t) \cdot V^n(t) \right\} \qquad (2)$$

The sequence of value functions $V^n$ produced by value iteration converges linearly to the optimal value function $V^*$. For some finite $n$, the actions that maximize Equation 2 form an optimal policy, and $V^n$ approximates its value. A commonly used stopping criterion specifies termination of the iteration procedure when

$$\|V^{n+1} - V^n\| < \frac{\epsilon(1 - \beta)}{2\beta} \qquad (3)$$

(where $\|X\| = \max\{|x| : x \in X\}$ denotes the supremum norm). This ensures that the resulting value function $V^{n+1}$ is within $\frac{\epsilon}{2}$ of the optimal function $V^*$ at any state, and that the resulting policy is $\epsilon$-optimal [19].

## 3  ADDs and MDPs

Algebraic decision diagrams (ADDs) [1] are a generalization of BDDs [10], a compact, efficiently manipulable data structure for representing boolean functions. These data structures have been used extensively in the VLSI CAD field and have enabled the solution of much larger problems than previously possible. In this section, we will describe these data structures and basic operations on them and show how they can be used for MDP representation.

### 3.1  Algebraic Decision Diagrams

A BDD represents a function $\mathcal{B}^n \to \mathcal{B}$ from $n$ boolean variables to a boolean result. Bryant [10] introduced the BDD in its current form, although the general ideas have been around for quite some time (e.g., as branching programs in the theoretical computer science literature). Conceptually, we can construct the BDD for a boolean function as follows. First, build a decision tree for the desired function, obeying the restrictions that along any path from root to leaf, no variable appears more than once, and that along every path from root to leaf, the variables always appear in the same order. Next, apply the following two reduction rules as much as possible: (1) merge any duplicate (same label and same children) nodes; and (2) if both child pointers of a node point to the same child, delete the node because it is redundant (with the parents of the node now pointing directly to the child of the node). The resulting directed, acyclic graph is the BDD for the function.[2] In practice, BDDs are generated and manipulated in the fully-reduced form, without ever building the decision tree.

ADDs generalize BDDs to represent real-valued functions $\mathcal{B}^n \to \mathcal{R}$; thus, in an ADD, we have multiple terminal nodes labeled with numeric values. More formally, an ADD denotes a function as follows:

1. The function of a terminal node is the constant function $f() = c$, where $c$ is the number labelling the terminal node.

2. The function of a nonterminal node labeled with boolean variable $X_1$ is given by

   $$f(x_1 \ldots x_n) = x_1 \cdot f_{then}(x_2 \ldots x_n) + \overline{x_1} \cdot f_{else}(x_2 \ldots x_n)$$

   where boolean values $x_i$ are viewed as 0 and 1, and $f_{then}$ and $f_{else}$ are the functions of the ADDs rooted at the *then* and *else* children of the node.

BDDs and ADDs have several useful properties. First, for a given variable ordering, each distinct function has a unique reduced representation. In addition, many common functions can be represented compactly because of isomorphic-subgraph sharing. Furthermore, efficient algorithms (e.g., depth-first search with a hash table to reuse previously computed results) exist for most common operations, such as addition, multiplication, and maximization. For example, Figure 1 shows a computation of the maximum of two ADDs. Finally, because BDDs and ADDs have been used extensively in other domains, very efficient implementations are readily available. As we will see, these properties make ADDs an ideal candidate to represent structured value functions in MDP solution algorithms.

---

[1] We ignore actions costs for ease of exposition. These impose no serious complications.

[2] We are describing the most common variety of BDD. Numerous variations exist in the literature.

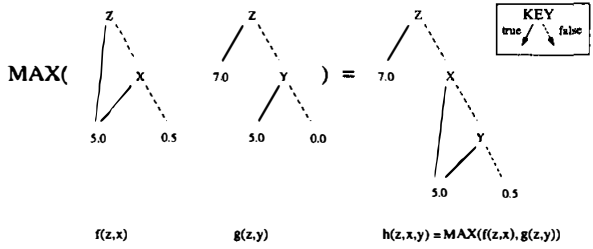

Figure 1: Simple ADD maximization example

### 3.2 ADD Representation of MDPs

We assume that the MDP state space is characterized by a set of variables $\mathbf{X} = \{X_1, \cdots, X_n\}$. Values of variable $X_i$ will be denoted in lowercase (e.g., $x_i$). We assume each $X_i$ is boolean, as required by the ADD formalism, though we discuss multi-valued variables in Section 5. Actions are often most naturally described as having an effect on specific variables under certain conditions, implicitly inducing state transitions. DBN action representations [13, 7] exploit this fact, specifying a local distribution over each variable describing the (probabilistic) impact an action has on that variable.

A DBN for action $a$ requires two sets of variables, one set $\mathbf{X} = \{X_1, \cdots, X_n\}$ referring to the state of the system before action $a$ has been executed, and $\mathbf{X}' = \{X'_1, \cdots, X'_n\}$ denoting the state after $a$ has been executed. Directed arcs from variables in $\mathbf{X}$ to variables in $\mathbf{X}'$ indicate direct causal influence and have the usual semantics [17, 13].[3] The conditional probability table (CPT) for each post-action variable $X'_i$ defines a conditional distribution $P^a_{X'_i}$ over $X'_i$—i.e., $a$'s effect on $X_i$—for each instantiation of its parents. This can be viewed as a function $P^a_{X'_i}(X_1 \ldots X_n)$, but where the function value (distribution) depends only on those $X_j$ that are parents of $X'_i$. No quantification is provided for pre-action variables $X_i$: since the process is fully observable, we need only use the DBN to predict state transitions. We require one DBN for each action $a \in \mathcal{A}$.

In order to illustrate our representation and algorithm, we introduce a simple adaptation of a process planning problem taken from [14]. The example involves a factory agent which has the task of connecting two objects $A$ and $B$. Figure 2(a) illustrates our representation for the action *bolt*, where the two parts are bolted together. We see that whether the parts are successfully connected, C, depends on a number of factors, but is independent of the state of variable P (*painted*). In contrast, whether part $A$ is punched, APU, after bolting depends only on whether it was punched before bolting.

Rather than the standard, locally exponential, tabular repre-sentation of CPTs, we use ADDs to capture regularities in the CPTs (i.e., to represent the functions $P^a_{X'_i}(X_1 \ldots X_n)$). This type of representation exploits context-specific independence in the distributions [9], and is related to the use of tree representations [7] and rule representations [18] of CPTs in DBNs. Figure 2(b) illustrates the ADD representation of the CPT for two variables, C' and APU'. While the distribution over C' is a function of its seven parent variables, this function exhibits considerable regularity, readily apparent by inspection of the table, which is exploited by the ADD. Specifically, the distribution over C' is given by the following formula:

$$P^{bolt}_{\mathbf{C}'}(\mathbf{C}, \mathbf{PL}, \mathbf{APU}, \mathbf{BPU}, \mathbf{ADR}, \mathbf{BDR}, \mathbf{BO}) =$$
$$[\mathbf{C} + \overline{\mathbf{C}}[(\mathbf{PL} \cdot \overline{\mathbf{APU}} + \overline{\mathbf{PL}}) \cdot \mathbf{ADR} \cdot \mathbf{BDR}$$
$$+ \mathbf{PL} \cdot \mathbf{APU} \cdot \mathbf{BPU}] \cdot \mathbf{BO}] \cdot 0.9$$

---

[3] We ignore the possibility of arcs among post-action variables, disallowing correlations in action effects. See [4] for a treatment of dynamic programming when such correlations exist.

(we ignore the zero entries). Similarly, the ADD for APU' corresponds to:

$$P^{bolt}_{\mathbf{APU}'}(\mathbf{APU}) = \mathbf{APU} \cdot 1.0$$

Reward functions can be represented similarly. Figure 2(c) shows the ADD representation of the reward function for this simple example: the agent is rewarded with 10 if the two objects are connected and painted, with a smaller reward of 5 when the two objects are connected but not painted, and is given no reward when the parts are not connected. The reward function, $R(X_1, \ldots, X_n)$, is simply

$$R(\mathbf{C}, \mathbf{P}) = \mathbf{C} \cdot \mathbf{P} \cdot 10.0 + \mathbf{C} \cdot \overline{\mathbf{P}} \cdot 5$$

This example action illustrates the type of structure that can be exploited by an ADD representation. Specifically, the CPT for C' clearly exhibits disjunctive structure, where a variety of distinct conditions *each* give rise to a specific probability of successfully connecting two parts. While this ADD has seven internal nodes and two leaves, a tree representation for the same CPT requires 11 internal nodes and 12 leaves. As we will see, this additional structure can be exploited in value iteration. Note also that the standard matrix representation of the CPT requires 128 parameters.

ADDs are often much more compact that trees when representing functions, but this is not always the case. The ordering requirement on ADDs means that certain functions can require an exponentially larger ADD representation than a well-chosen tree; similarly, ADDs can be exponentially smaller than decision trees. Our initial results suggest that such pathological examples are unlikely to arise in most problem domains (see Section 5), and that ADDs offer an advantage over decision trees.

## 4 Value Iteration using ADDs

In this section, we present an algorithm for optimal policy construction that avoids the explicit enumeration of the state space. SPUDD (stochastic planning using decision diagrams) implements classical value iteration, but uses ADDs to represent value functions and CPTs. It exploits the regularities in the action and reward networks, made explicit by the ADD representation described in the previous section, to discover regularities in the value functions it constructs. This often yields substantial savings in both space and computational time. We first introduce the algorithm in a conceptually clear way, and then describe certain optimizations.

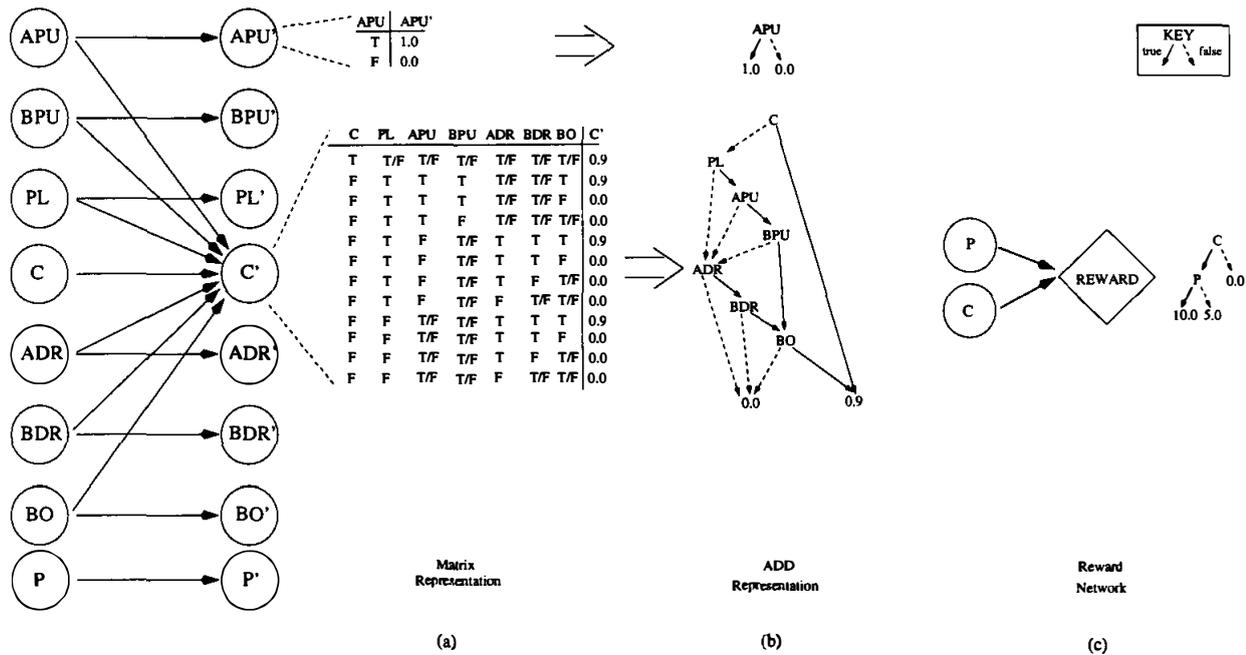

Figure 2: Small FACTORY example: (a) action network for action *bolt*; (b) ADD representation of CPTs (action diagrams); and (c) immediate reward network and ADD representation of the reward table.

OBDDs have been explored in previous work in AI planning [11], where universal plans (much like policies) are generated for nondeterministic domains. The motivation in that work, avoiding the combinatorial explosion associated with state space enumeration, is similar to ours; but the details of the algorithms, and how the representation is used to represent planning domains, is quite different.

### 4.1 The Basic SPUDD Algorithm

The SPUDD algorithm, shown in Figure 3, implements a form of value iteration, producing a sequence of value functions $V^0, V^1, \cdots$ until the termination condition is met. Each $i$ stage-to-go value function is represented as an ADD denoted $V^i(X_1, \ldots, X_n)$. Since $V^0 = R$, the first value function has an obvious ADD representation. The key insight underlying SPUDD is to exploit the ADD structure of $V^i$ and the MDP representation itself to discover the appropriate ADD structure for $V^{i+1}$. Expected value calculations and maximizations are then performed at each terminal node of the new ADD rather than at each state.

Given an ADD for $V^i$, Step 3 of SPUDD produces $V^{i+1}$. When computing $V^{i+1}$, the function $V^i$ is viewed as representing values at *future* states, after a suitable action has been performed with $i+1$ stages remaining. So variables in $V^i$ are first replaced by their *primed*, or post-action, counterparts (Step 3(a)), referring to the state with $i$ stages-to-go; this prevents them from being confused with unprimed variables that refer to the state with $i + 1$ stages-to-go. Figure 4(a) shows the zero stage-to-go primed value diagram, $V'^0$, for our simple example.

For each action $a$, we then compute an ADD representation of the function $V_a^{i+1}$, denoting the expected value of performing action $a$ with $i + 1$ stages to go given that $V^i$ dictates $i$ stage-to-go value. This requires several steps, described below. First, we note that the ADD-represented functions $P_{X_i'}^a$, taken from the action network for $a$, give the (conditional) probabilities that variables $X_i'$ are made *true* by action $a$. To fit within the ADD framework, we introduce the *negative action diagrams*

$$\overline{P_{X_i'}^a}(X_1, \ldots, X_n) = (1 - P_{X_i'}^a(X_1, \ldots, X_n))$$

which gives the probability that $a$ will make $X_i'$ false. We then define the *dual action diagrams* $Q_{X_i'}^a$ as the ADD rooted at $X_i'$, whose *true* branch is the action diagram $P_{X_i'}^a$ and whose *false* branch is the negative action diagram $\overline{P_{X_i'}^a}$:

$$Q_{X_i'}^a(X_i'; X_1, \ldots X_n) = X_i' \cdot P_{X_i'}^a(X_1, \ldots X_n) + \overline{X_i'} \cdot \overline{P_{X_i'}^a}(X_1, \ldots X_n)) \quad (4)$$

Intuitively, $Q_{X_i'}^a(x_i'; x_1, \ldots x_n)$ denotes $P(X_i' = x_i' | X_1 = x_1, \cdots, X_n = x_n)$ (under action $a$). Figure 4(a) shows the dual action diagram for the variable C' from the example in Figure 2(b).

In order to generate $V_a^{i+1}$, we must, for each state $s$, combine the $i$ stage-to-go value for each state $t$ with the probability of reaching $t$ from $s$. We do this by multiplying, in turn, the dual action diagrams for each variable $X_j'$ by $V'^i$

1. Set $V^0 = R$ where $R$ is the immediate reward diagram; set $i = 0$

2. Create *dual* action diagrams, $Q^a_{X'_i}(X'_i, X_1, \ldots X_n)$ for each $a \in \mathcal{A}$, and for each $X'_i \in \mathbf{X'}$

3. Repeat until $\|V^{i+1} - V^i\| < \frac{\epsilon(1-\beta)}{2\beta}$

   (a) Swap all variables in $X^i$ with *primed* versions to create $X'^i$
   
   (b) For all $a \in \mathcal{A}$
       Set $temp = V'^i$
       For all *primed variables*, $X'_j$ in $V'^i$
          $temp = temp * Q^a_{X'_j}$
          Set $temp$ = Sum the sub-diagrams of $temp$ over the *primed* variable $X'_j$
       End For
       Multiply the result by discounting factor $\beta$ and add $R$ to obtain $V^i_a$
       End For
   
   (c) Maximize over all $V^i_a$'s to create $V^{i+1}$.
   
   (d) Increment i
   
   End Repeat

4. Perform one more iteration and assign to each terminal node the actions $a$ which contributed the value in the value ADD at that node; this yields the $\epsilon$-*optimal policy* ADD, $\pi^*$. Note that terminal nodes which have the same values for multiple actions are assigned all possible actions in $\pi^*$.

5. Return the value diagram $V^{i+1}$ and the optimal policy $\pi^*$.

Figure 3: SPUDD algorithm

and then eliminating $X'_j$ by summing over its values in the resultant ADD. More precisely, by multiplying $Q^a_{X'_j}$ by $V'^i$, we obtain a function $f(X'_1, \cdots, X'_n, X_1, \cdots X_n)$ where

$$f(x'_1, \cdots, x'_n, x_1, \cdots x_n) = V'^i(x'_1, \cdots, x'_n) P(x'_j | x_1, \ldots x_n)$$

(assuming transitions induced by action $a$). This intermediate calculation is illustrated in Figure 4(b), where the dual diagram for variable C' is the first to be multiplied by $V'^0$. Note that C' lies at the root of this ADD. Once this function $f$ is obtained, we can eliminate dependence of future value on the specific value of $X'_j$ by taking an expectation over both of its truth values. This is done by summing the left and right subgraphs of the ADD for $f$, leaving us with the function

$$g(X'_1, \cdots, X'_{j-1}, X'_{j+1}, \cdots, X'_n, X_1, \ldots X_n) = \sum_{x'_j} V'^i(X'_1, \cdots, x'_j, \cdots, X'_n) P(x'_j | X_1, \ldots X_n)$$

This is illustrated in Figure 4(c), where the variable C' is eliminated. This ADD denotes the expected *future* value (or 0 stage-to-go value) as a function of the parents of C' with 1 stage-to-go and all post-action variables except C' with 0 stages-to-go.

This process is repeated for each post-action variable $X'_j$ that occurs in the ADD for $V'^i$: we first multiply $Q^a_{X'_j}$ into the intermediate value ADD, then eliminate that variable by taking an expectation over its values. Once all primed variables have been eliminated, we are left with a function

$$h(X_1, \cdots, X_n) = \sum_{x'_1, \cdots, x'_n} V'^i(x'_1, \cdots, x'_n) P(x'_1 | X_1, \ldots X_n) \cdots P(x'_n | X_1, \ldots X_n)$$

By the independence assumptions embodied in the action network, this is precisely the expected future value of performing action $a$. By adding the reward ADD $R$ to this function, we obtain an ADD representation of $V^{i+1}_a$. Figure 5 shows the result for our simple example. The remaining primed variable P' in Figure 4(c) has been removed, producing $V^1_{\text{bolt}}$ using a discount factor of 0.9. Finally, we take the maximum over all actions to produce the $V^{i+1}$ diagram. Given ADDs for each $V^{i+1}_a$, this requires simply that one construct the ADD representing $\max_{a \in \mathcal{A}} V^{i+1}_a$.

The stopping criterion in Equation 3 is implemented by comparing each pair of successive ADDs, $V^{i+1}$ and $V^i$. Once the value function has converged, the $\epsilon$-optimal policy, or *policy ADD*, is extracted by performing one further dynamic programming backup, and assigning to each terminal node the actions which produced the maximimizing value. Since each terminal node represents some state set of states $S$, the set of actions thus determined are each optimal for any $s \in S$.

### 4.2 Optimizations

The algorithm as described in the last section, and as shown in Figure 3, suffers from certain practical difficulties which make it necessary to introduce various optimizations in order to improve efficiency with respect to both space and time. The problems arise in Step 3(b) when $V'^i$ is multiplied by the dual action diagrams $Q^a$. Since there are potentially $n$ primed variables in the ADD for $V'^i$ and $n$ unprimed variables in the ADD for $Q^a$, there is an intermediate step in which a diagram is created with (potentially) up to $2n$ variables. Although this will not be the case in general, it was deemed necessary to modify the method in order to deal with the possibility of this problem arising. Furthermore, a large computational overhead is introduced by re-calculating the joint probability distributions over the primed variables at each iteration. In this section, we first discuss optimizations for dealing with space, followed by a method for optimizing computation time.

The increase in the diagram size during Step 3(b) of the algorithm can be countered by approaching the multiplications and sums slightly differently. Instead of blindly multiplying the $V'^i$ by the dual action diagram for the variable at the root of $V'^i$, we can traverse the ADD for $V'^i$ to the level of the last variable in the ADD ordering, then multiply and sum the sub-diagrams rooted at this variable by the corresponding dual diagram. This process will only remove the dependency of the $V'^i$ on a primed variable for a given branch, and will therefore only introduce a single diagram of $n$ unprimed variables at a leaf node of $V'^i$. By recursively carrying out this procedure using the structure of the ADD for $V'^i$, the intermediate stages never grow too large. Essentially, the additional unprimed variables are introduced only at specific points in the ADD and the corresponding primed variable immediately eliminated—this is much like the tree-structured dynamic programming algorithm of [7].

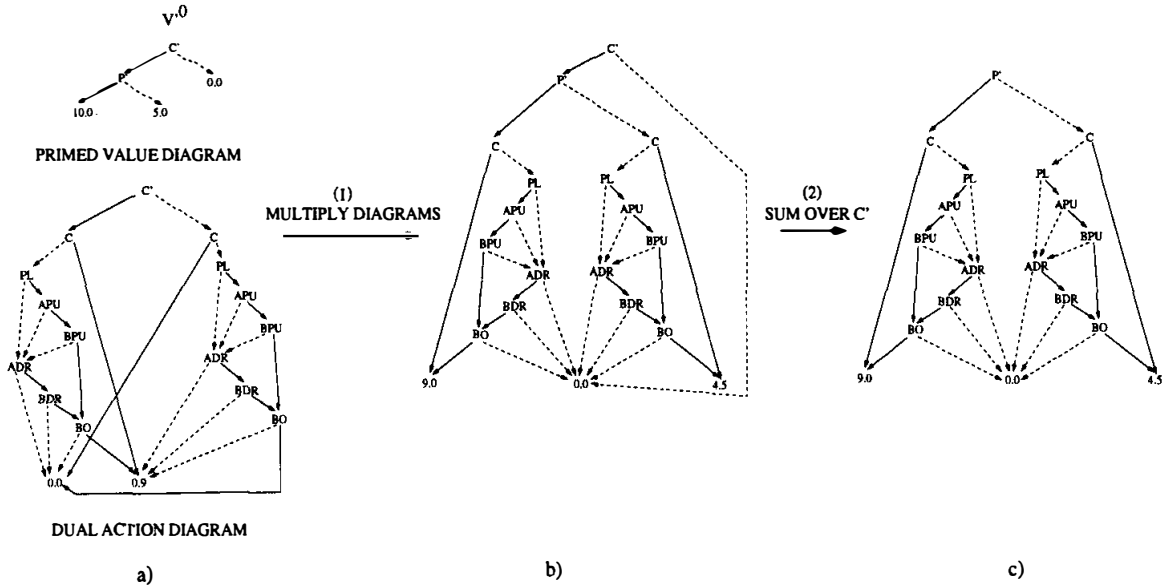

Figure 4: First Bellman backup for the *Value Iteration using ADDs* algorithm. (a) *0-stage-to-go primed* value diagram, and dual action diagram for variable C', $Q^a_{C'}$. (b) Intermediate result after multiplying $V'^0$ with $Q^a_{C'}$. (c) Intermediate result after quantifying over C'.

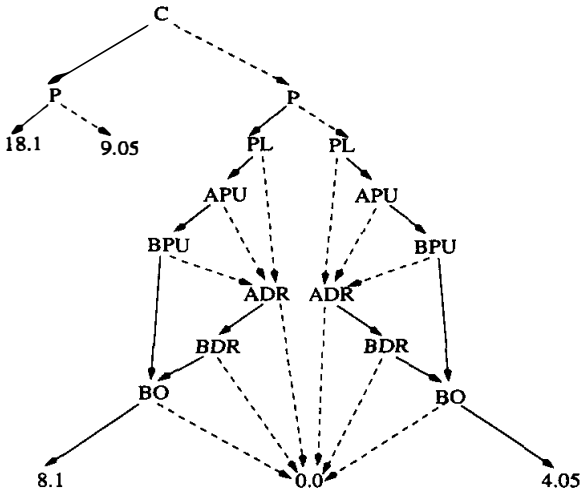

Figure 5: Resulting *1-stage-to-go* value diagram for action *bolt*, $V^1_{bolt}$.

Unfortunately, this method requires a great deal of unnecessary, repeated computation. Since the action diagrams for a given problem do not change during the generation of a policy, the joint probability distribution $Pr(s, a, t)$ from Equation 2 could be pre-computed. In our case, this means we could take the product of all dual action diagrams for a given action $a$, as shown in Equation 5 below, prior to a specific value iteration. We refer to this product diagram, $P^a$, as the *complete* action diagram for action $a$:

$$P^a(X'_1, \ldots, X'_n, X_1, \ldots, X_n) = \prod_{i=1}^{n} Q^a_{X'_i}(X'_i; X_1, \ldots, X_n)$$

(5)

The resulting function $P^a$ provides a representation of the state transition probabilities for action $a$. This explicit $P^a$ function could then be multiplied by the $V'^n$ during Step 3 of the algorithm, and then primed variables eliminated. Although this may lead to a substantial savings in computation time, it will again generate diagrams with up to $2n$ variables.

As a compromise, we implemented a method where the space-time trade-off can be addressed explicitly. A "tuning knob" enables the user to find a middle ground between the two methods mentioned above. We accomplish this by pre-computing only subsets of the complete action diagram. That is, we break the large diagram up into a few smaller pieces. The set of variables $(X_1, \ldots, X_n)$ is divided into $m$ subsets, preserving the total ordering (e.g., $[X_1, \ldots, X_{i_1}], [X_{i_1+1}, \ldots, X_{i_2}], \ldots, [X_{i_m}, \ldots X_n]$), and the complete action diagrams are pre-computed for each subset (e.g., $(P^a(X'_{i_j}, \ldots, X'_{i_{j+1}}, v_1, \ldots, X_n))$. Step 3(b) of the algorithm must be modified as shown in Figure 6. The primed value diagram $V'^i$ is traversed to the top of the second level $(i_1 + 1)$, and the procedure is carried out recursively on each sub-diagram rooted at variables $X'_{i_1+1}$. If a level is reached with no variables below it, then the sub-diagram rooted at each variable $X'_{i_m}$ of $V'^i$ is multiplied with the corresponding subset of the complete action diagram, $P^a(X'_{i_m},\ldots,X'_n,X_1,\ldots,X_n)$, and summed over primed variables $X'_k$, $k > i_m$. In this way, the diagrams are kept small by making sure that enough elimination occurs to balance the effects of multiplying by complete action diagrams. The space and time requirements can then be controlled by the number of subsets the complete action diagrams are broken into. In theory, the more subsets, the smaller the space requirements and the larger the time requirements. Although we have been able to produce substantial changes in the space and time requirements of the algorithm using this tuning knob, its effects are still unclear. At present, we choose the $m$ subsets of variables by simply building the complete action diagrams according to some variable ordering until they reach a user-defined size limit, at which point we start on the next subset. We note that this space-time tradeoff bears some resemblance to the space-time tradeoffs that arise in probabilistic inference algorithms like variable elimination [15].

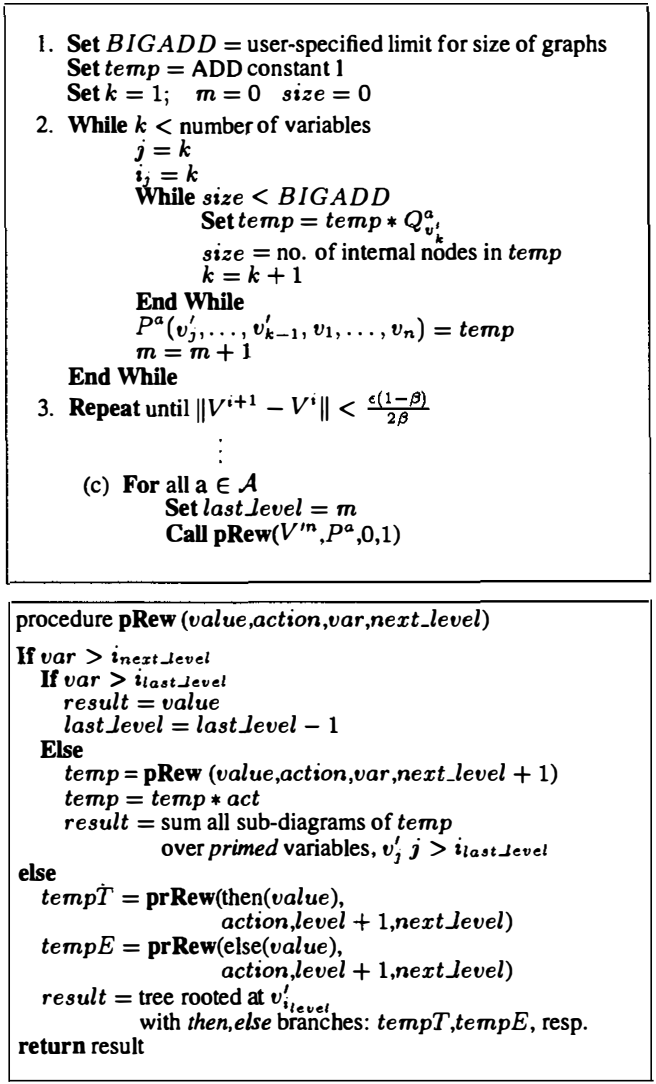

Figure 6: Modified SPUDD algorithm

Although we have not implemented heuristics for variable ordering, there are some simple ordering methods that could improve space efficiency. For instance, if we order variables so that primed variables with many shared parents are eliminated together, the number of unprimed variables introduced will be kept relatively small relative to the number of primed variables eliminated. More importantly, we must develop more refined heuristics that keep the ADDs small rather than minimizing the number of variables introduced.

This revised procedure (Figure 6) has a small inefficiency, as our results in the next section will show. Since we are pre-computing subsets of the complete action diagrams, any variables which are included in the domain, but are not relevant to its solution, will be included in these pre-computed diagrams. This will increase the size of the intermediate representations and will add overhead in computation time. It is important to be able to discard them, and to only compute the policy over variables that are relevant to the value function and policy [7]. A possible way to deal with these types of variables in our algorithm would be to progressively build the complete action diagrams during the iterative procedure. In this way, only the variables relevant to the domain would be added.

## 5 Data and Results

The procedure described above was implemented using the CUDD package [20], a library of C routines which provides support for manipulation of ADDs. Experimental results described in this section were all obtained using a dual-processor *SUN SPARC Ultra 60* running at 300Mhz with 1 Gb of RAM, with only a single processor being used. The SPUDD algorithm was tested on three different types of examples, each type having MDP instances with different numbers of variables, hence a wide variety of state space sizes. The first example class consists of various adaptations of a process planning problem taken from [14]. The second and third example classes consist of synthetic problems taken from [7, 8]. These are designed to test best- and worst-case behavior of SPUDD.[4]

The first example class consists of process planning problems taken from [14], involving a factory agent which must paint two objects and connect them. The objects must be smoothed, shaped and polished and possibly drilled before painting, each of which actions require a number of tools which are possibly available. Various painting and connection methods are represented, each having an effect on the quality of the job, and each requiring tools. The final product is rewarded according to what kind of quality is needed. Rewards range from 0 to 10 and a discounting factor of 0.9 was used throughout.

The examples used here, unlike the one described in Section 3, were not designed with any structure in mind which could be taken advantage of by an ADD representation. In the original problem specification, three ternary variables were used to represent painting quality of each object (*good, poor* or *false*), and the connection quality (*good, bad* or *false*). However, as discussed above, ADDs can only rep-resent binary variables, so that each ternary variable was expanded into two binary ones. For example, the variable *connected*, describing the type of connection between the two objects, was represented by boolean variables *connected* and *connected_well*. This expansion enlarges the state space by a factor of 4/3 for each ternary variable so expanded (by introducing unreachable states). A number of FACTORY examples were devised, with state space sizes ranging from 55 thousand to 268 million.

---
[4]Data for these problems can be found at the Web page: www.cs.ubc.ca/spider/staubin/Spudd/index.html.

Optimal policies were generated using SPUDD and a *structured policy iteration* (SPI) implementation for comparison purposes [7]. Results, displayed in Table 1, are presented for SPUDD running on six FACTORY examples, and for SPI running on five. SPI was not run on the *factory4* example, because its estimated time and space requirements exceeded available capacity. SPI implements modified policy iteration using trees to represent CPTs and intermediate value and policy functions. SPI, however, does allow multi-valued variables—so versions of each example were tested in SPI using both ternary variables, and thier binary expansion. Table 1 shows the number of ternary variables in each example, along with the total number of variables. The state space sizes of each FACTORY example are shown for both the original and the binary-expansion formulations. SPUDD was only run on the binary-expanded versions.

The examples labelled *factory1* and *factory2* differ only by a single binary variable, which is not affected by any action in the domain, and which does not itself affect any other variables. Hence, the number of internal nodes resulting in Table 1 are identical for the two examples. This variable was added in order to show how structured representations like SPUDD and SPI can effectively discard variables which do not affect the problem at hand, as discussed in Section 4.2. Since SPUDD pre-computes the *complete* action diagrams, as shown in Figure 6, the running time for SPUDD almost doubles when this new variable is added, since it creates overhead for the iterative procedure. This problem could be circumvented using the method described at the end of Section 4.2.

Running times are shown for SPUDD and SPI. However, the algorithms do not lend themselves easily to comparisons of running times, since implementation details cloud the results; so running times will not be discussed further here. The SPI results are shown in order to compare the sizes of the final value function representations, which give an indication of complexity for policy generation algorithms. However, a question arises when comparing such numbers about the variable orderings, as mentioned in Section 3. The variable ordering for SPUDD is chosen prior to runtime and remains the same during the entire process. No special techniques were used to choose the ordering, although it may be argued that good orderings could be gleaned from the MDP specification. Variable orderings within the branches of the tree structure in the SPI algorithm are determined primarily by the choice of ordering in the reward function and action descriptions [7]. Again, no special techniques were used to choose the variable ordering in SPI. Finding the optimal variable orderings in either case is a difficult problem, and we assume here that neither algorithm has an advantage in this regard. Dynamic reordering algorithms are available in CUDD, and have been implemented but not yet fully tested in SPUDD (see below).

In order to compare representation sizes, we compare the number of internal nodes in the value function representations only. This is most important when doing dynamic programming back-up steps and is a large factor in determining both running time and space requirements. Furthermore, we compare numbers from SPUDD using binary representations with numbers from SPI using binary/ternary representations in order not to disadvantage SPI, which can make use of ternary variables. We also compare both implementations using only binary variables. The *equivalent tree leaves* column in Table 1 gives the number of leaves of the totally ordered binary tree (and hence the number of internal nodes) that results in expanding the value ADD generated by SPUDD. These numbers give the size of a tree that would be generated if a total ordering was imposed. Comparing these numbers with the numbers generated by SPI give an indication of the savings that occur due to the relaxation of the total ordering constraint. The rightmost column in Table 1 shows the ratio of the number of internal nodes in the tree representation to the number in the ADD representation. We see that reductions of up to 30 times are possible, when comparing only binary representations to binary/ternary representations, and reductions of over 40 times when comparing the same binary representations. These space savings also showed up in the amount of memory used. For example, the *factory3* example took 691Mb of memory using SPI, and only 148Mb using SPUDD. The *factory4* example took 378Mb of space using SPUDD.

The BIGADD limit (see Figure 6) was set to 10000 for the *factory*, *factory0*, *factory1* and *factory2* examples and to 20000 in the *factory3* and *factory4* examples. These limits broke up the complete action diagrams into $m = 2$ or 3 pieces, with typically 6000-10000 nodes in the first and second and under 1000 nodes in the third if it existed. In the large examples (*factory2, 3 and 4*), it was not possible (with 1Gb of RAM) to generate the full complete action diagram ($m = 1$), and running times became too large when BIGADD was set to 1. The functionality of this "tuning knob" was not fully investigated, but, along with studies of different heuristics for variable grouping, is an interesting avenue for future exploration.

For comparison purposes, *flat* (unstructured) value iteration was run on both the *factory* and *factory0* examples. The times taken for these problems were 895 and 4579 seconds, respectively. For the larger problems, memory limitations precluded completion of the flat algorithm.

In order to examine the worst-case behaviour, we tested SPUDD on a series of examples, drawn from [7, 8], in which every state has a unique value; hence, the ADD representing the value function will have a number of terminal nodes exponential in the number of state variables. The problem EXPON involves $n$ ordered propositions and $n$ actions, one for each proposition. Each action makes its corresponding proposition true, but causes all propositions lower in the order to become false. A reward is given only if all variables are true. The problem is representable in $O(n^2)$ space using ADDs; but the optimal policy winds through the entire state space like a binary counter. This problem causes worst-case behaviour for SPUDD because all $2^n$ states have different values. SPUDD was tested on the EXPON example with 6, 8, 10 and 12 variables, leading to state spaces with sizes 64, 256, 1024 and 4096, respectively. The initial reward and the discounting factor in these examples must be scaled to accommodate the $2^n$-step lookahead for the largest problem (12 variables), and were set to $10^{16}$ and 0.99, respectively.[5] Figure 7 compares the running times of SPUDD and (flat) value iteration plotted (in log scale) as a function of the number of variables. Running times for both algorithms exhibit exponential growth with the number of variables, as expected.[6] It is not surprising that flat value iteration performs better in this type of problem since there is absolutely no structure that can be exploited by SPUDD. However, the overhead involved with creating ADDs is not overly severe, and tends to diminish as the problems grow larger. With $n = 12$, SPUDD takes less than 10 times longer than value iteration.

| Example Name | State space size | | | SPUDD - Value | | | | SPI - Value | | | ratio of tree nodes: ADD nodes |
|---|---|---|---|---|---|---|---|---|---|---|---|
| | variables ternary | total | states | time (s) | internal nodes | leaves | equiv. tree leaves | time (s) | internal nodes | leaves | |
| factory | 3<br>0 | 14<br>17 | 55296<br>131072 | -<br>78.0 | -<br>828 | -<br>147 | -<br>8937 | 2210.6<br>2188.23 | 6721<br>9513 | 7879<br>9514 | 8.12<br>11.48 |
| factory0 | 3<br>0 | 16<br>19 | 221184<br>524288 | -<br>111.4 | -<br>1137 | -<br>147 | -<br>14888 | 5763.1<br>6238.4 | 15794<br>22611 | 18451<br>22612 | 13.89<br>19.89 |
| factory1 | 3<br>0 | 18<br>21 | 884736<br>2097132 | -<br>279.0 | -<br>2169 | -<br>178 | -<br>49558 | 14731.9<br>15430.6 | 31676<br>44304 | 37315<br>44305 | 14.60<br>20.43 |
| factory2 | 3<br>0 | 19<br>22 | 1769472<br>4194304 | -<br>462.1 | -<br>2169 | -<br>178 | -<br>49558 | 14742.4<br>15465.0 | 31676<br>44304 | 37315<br>44305 | 14.60<br>20.43 |
| factory3 | 4<br>0 | 21<br>25 | 10616832<br>33554432 | -<br>3609.4 | -<br>4711 | -<br>208 | -<br>242840 | 98340.0<br>112760.1 | 138056<br>193318 | 168207<br>193319 | 29.31<br>41.04 |
| factory4 | 4<br>0 | 24<br>28 | 63700992<br>268435456 | -<br>14651.5 | -<br>7431 | -<br>238 | -<br>707890 | -<br>- | -<br>- | -<br>- | -<br>- |

Table 1: Results for FACTORY examples.

One can similarly construct a "best-case" series of examples, where the value function grows linearly in the number of problem variables. Specifically, the problem LINEAR involves $n$ variables and has $n+1$ distinct values. The MDP can be represented in $O(n^2)$ space using ADDs and the optimal value function can be represented in $O(n)$ space with an ADD (see [8] for further details).[7] Hence, the inherent structure of such a problem can easily be exploited. As seen in Figure 8, SPUDD clearly takes advantage of the structure in the problem, as its running time increases linearly with the number of variables, compared to an exponential

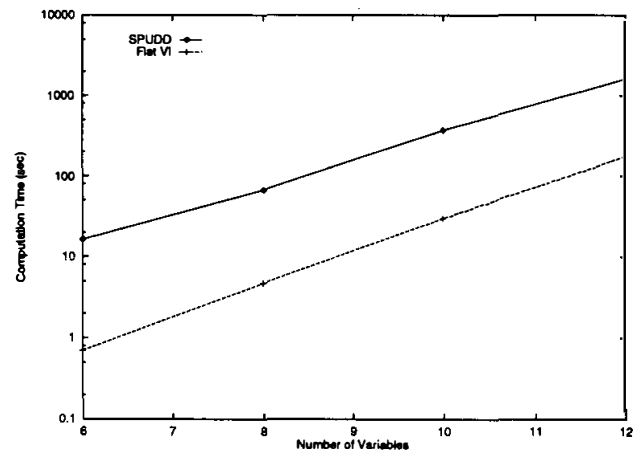

Figure 7: Worst-case behavior for SPUDD.

increase in running time associated with flat value iteration.

## 6 Concluding Remarks

In this paper, we described SPUDD, an implementation of value iteration, for solving MDPs using ADDs. The ADD representation captures some regularities in system dynamics, reward and value, thus yielding a simple and efficient representation of the planning problem. By using such a compact representation, we are able to solve certain types of problems that cannot be dealt with using current techniques, including explicit matrix and decision tree methods. Though the technique described in this paper has not yet been tested extensively on realistic domains, our preliminary results are encouraging.

One drawback of using ADDs is the requirement that variables be boolean. Any (finite-valued) non-boolean variable can be split into a number of boolean variables, generally in a way that preserves at least some of the structure of the original problem (see above), though it often makes the new state space larger than the original. Conceptually, there is no difficulty in allowing ADDs to deal with multi-valued variables (all algorithms and canonicity results carry over easily). However, for domains with relatively few multi-valued variables, SPUDD does not appear to be handicapped by the requirement of variable splitting.

---

[5] Since the value obtained at the state furthest from the goal is the goal reward discounted by the number of system states (since each must be visited along the way), the goal reward must be set very high to ensure that the value at this state is not (practically) zero.

[6] The running times are especially large due to the nature of the problem which requires a large number of iterations of alue iteration to converge.

[7] Of course, *best-case* behavior for SPUDD involves a problem in which all variables are irrelevant to the value function. This problem represents a "best case" in which all variables are required in the prediction of state value.

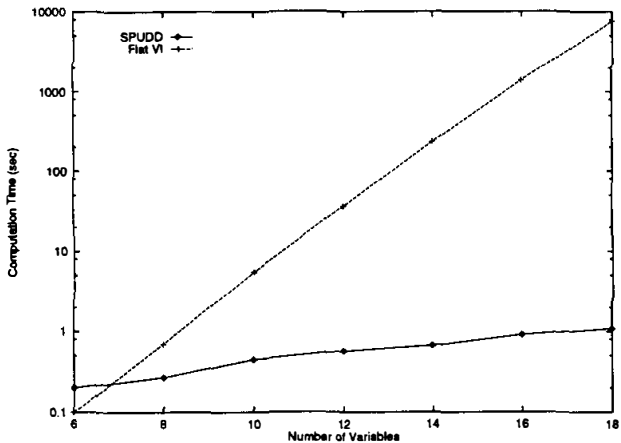

Figure 8: Best-case behavior for SPUDD.

At present, SPUDD uses a static user-defined variable ordering in order not to cloud the initial results with the effects of dynamic variable reordering. However, dynamic reordering of the variables at runtime can make significant improvements in both the space required, by finding a more compact representation, and in the running time, by choosing more appropriate subsets of variables as discussed in Section 4.2. The *CUDD* package provides a rich set of dynamic reordering algorithms [20]. Typically, when the ADD grows too large, variable reorderings are attempted by following one of these algorithms, and a new ordering is chosen which minimizes the space needed. Some of the available techniques are slight variations of existing techniques while some others were specifically developed for the package. It may be necessary, however, to implement a new heuristic which takes into account the variable subsets which influence the running time. Future work will include more complete experimentation with automatic dynamic reordering in SPUDD. Another extension of SPUDD would be the implementation of other dynamic programming algorithms, such as modified policy iteration, which are generally considered to converge more quickly than value iteration in practice. Finally, we hope to explore approximation methods within the ADD framework, such as have previously been researched in the context of decision trees [6].

## Acknowledgements

Thanks to Richard Dearden for helpful comments and for providing both his SPI code and example descriptions for comparison purposes. St-Aubin was supported by NSERC. Hu was supported by NSERC. Boutilier was supported by NSERC Research Grant OGP0121843 and IRIS-III Project "Dealing with Actions."